\documentclass[11pt]{article}

\usepackage[english]{babel}
\usepackage[letterpaper,margin=1in]{geometry} 

\usepackage{amsmath}
\usepackage{amssymb}
\usepackage{graphicx}
\usepackage{eqnarray}
\usepackage[colorlinks=true, allcolors=blue]{hyperref}
\usepackage{array}
\newcolumntype{P}[1]{>{\centering\arraybackslash}p{#1}}
\newcolumntype{M}[1]{>{\centering\arraybackslash}m{#1}}

\usepackage{lscape}

\newcommand{\mnamelong}{Bidirectional Recurrent Attention for Time series Imputation}
\newcommand{\mname}{BRATI}

\title{BRATI: Bidirectional Recurrent Attention for Time-Series Imputation}

\author{
    Armando Collado-Villaverde\textsuperscript{1}\thanks{\texttt{armando.collado@uah.es}} \and
    Pablo Mu\~noz\textsuperscript{1}\thanks{\texttt{pablo.munoz@uah.es}} \and
    Mar\'ia D. R-Moreno\textsuperscript{1}\thanks{Corresponding author. \texttt{malola.rmoreno@uah.es}}
}
\date{}

\begin{document}

\maketitle

\begin{abstract}
Missing data in time-series analysis poses significant challenges, affecting the reliability of downstream applications. Imputation, the process of estimating missing values, has emerged as a key solution. This paper introduces BRATI, a novel Deep Learning model designed to address multivariate time-series imputation by combining Bidirectional Recurrent Networks and Attention mechanisms. BRATI processes temporal dependencies and feature correlations across long and short time horizons, utilizing two imputation blocks that operate in opposite temporal directions. Each block integrates recurrent layers and attention mechanisms to effectively resolve long-term dependencies.

We evaluate BRATI on three real-world datasets under diverse missing-data scenarios: randomly missing values, fixed-length missing sequences, and variable-length missing sequences. Our findings demonstrate that BRATI consistently outperforms state-of-the-art models, delivering superior accuracy and robustness in imputing multivariate time-series data.

\end{abstract}

\section{Introduction}
Multivariate time-series data are used in a wide range of applications, such as economics \cite{HSIEH20112510, pmlr-v48-bauer16}, meteorology \cite{NIPS2015_07563a3f, yi2016st-mvl, Zhang2020}, transportation \cite{Lai2018, Celikoglu2007, Karlaftis2011, 7929980, Zhang2017}, health-care \cite{NEURIPS2018_934b5358, Kaushik2020, Melin2020, PhysioNet} or predictive maintenance \cite{9131796, Abbasi_2019, 8342870, Barrero2021}. These data are often employed for classification and regression tasks but they frequently present missing values due to unexpected reasons, such as malfunctions in measurement devices, communication errors, or cost-saving strategies, especially in the healthcare domain. The presence of missing values can significantly degrade the performance of downstream models, making it essential to handle missing data effectively.

There are two common approaches to deal with missing values. The first one is deletion, which consists of removing samples that contain any missing values. Nevertheless, this practice makes the data incomplete and can destroy the time relationship of the data, especially when the missing rate is high \cite{Che2018}. Additionally, removing samples with any missing values leads to biased predictions by the downstream models. The second approach is to impute the missing values, that is, to estimate the missing data using the observed values. This option offers some critical advantages over deletion. Partially observed data can still contain useful information that otherwise would be discarded. Moreover, correctly imputed values can also remove the bias that could be introduced by deletion. Furthermore, on some cases, the missing values and patterns provide useful information and are noted as informative missingness \cite{RUBIN1976}.

\begin{figure}
\centering
\includegraphics[width=0.6\textwidth]{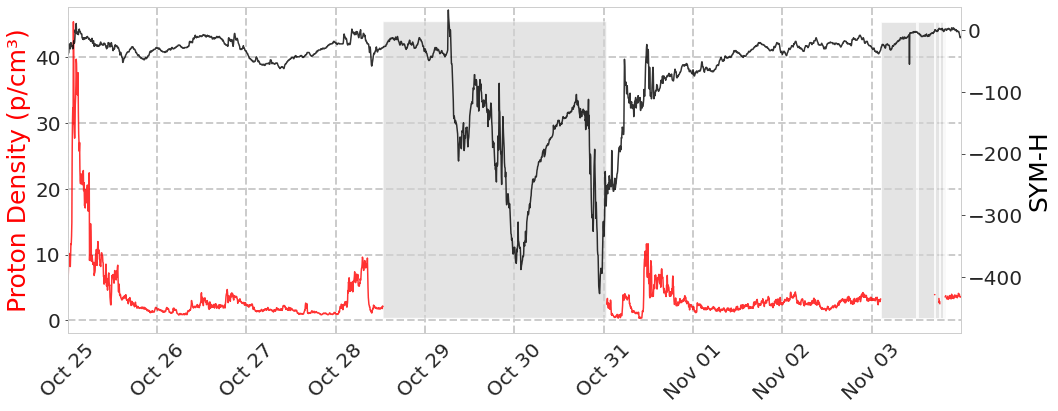}
\caption{\label{fig:storm}Example of the missing plasma (red line) values (shaded in gray) during a geomagnetic storm, when the intensity of the storm reaches its peak (represented by the geomagnetic index SYM-H, in black).}
\end{figure}

Missing data can be classified into three groups depending on its nature. First, \textbf{Missing Completely At Random (MCAR)}; in this case, the events that lead to missing values are independent of both the observable and unobservable variables and occur entirely at random \cite{polit2008nursing}. Second, \textbf{Missing At Random (MAR)}; the missing values depend on the observed values \cite{Heitjan1996}. Third, \textbf{Missing Not At Random (MNAR)}; the missing values depend on both the observed and unobserved variables \cite{gomer2021subtypes}, that is, the value of the variable that is missing is related to the reason it is missing. Most of the literature is focused on the MCAR case. However, in many real-world datasets, the missing values appear not in a random manner, as they are usually related to the reason they are missing. For example, on predictive maintenance datasets, such as wind farm turbines, missing data can appear when the wind is too strong and the power is limited \cite{notatrandomwind}. One dataset that we evaluate in this work is the geomagnetic storms solar wind observations, which presents missing values in an MNAR manner, due to how the instruments that collect the data work: during intense solar storms those instruments get saturated and stop providing valid measurements \cite{plasma-saturation}, being an example shown in Fig. \ref{fig:storm}. Another common example is found in the health-related datasets, when some measurements are only collected when the status of the patient changes and specific tests are required \cite{saits2023}, resulting in a sequence of missing values in between the tests.

Given the important nature of the MNAR values, we have decided to evaluate the performance of the baseline models and our proposed one on both cases, when the values are MCAR and when the values are missing in patterns, such as sequences of several values missing in sequence, which is common outside the MCAR case.

Nevertheless, the advantages of the imputation directly rely on its effectiveness, since a bad imputation can be even more harmful than deletion. A lot of prior work has been done toward filling the missing values with statistical approaches. However, most of them require strong assumptions on the missing values, such as imputing using the mean or median values, or using the k-nearest neighbors \cite{Acua2004}. Other methods are essentially linear, limiting their capabilities, such as linear regression, spline interpolation \cite{splines} or ARMA and ARIMA based models \cite{Ansley1984}. Kreindler et al. \cite{Kreindler2006-og} proposed the smoothing of the missing values, assuming that there is little to no variability during the missing period and the imputation can be done by smoothing over nearby values. However, those methods are unable to capture complex patterns or correlations between multiple variables. Due to those limitations, other imputation methods were developed to impute missing data with greater accuracy, such as kernel methods, matrix completion and factorization \cite{Koren2009, mazumder2010spectral, Koren22009}.

Recently, the literature has increasingly focused on the use of Deep Learning (DL) models for imputation tasks, achieving significant results \cite{wang2024deep}. Various approaches have been proposed for time-series data \cite{qi2020imitative, ma2019end, saits2023, zhang2019attain, cao2018brits, luo2018multivariate, che2018recurrent} and non-time-series data \cite{Kim2020-rp, Raad2022, Lee_2019_CVPR, mouselinos2021main}. Models like STING \cite{oh2021sting}, which utilize GANs combined with bidirectional RNNs and innovative attention mechanisms, and frameworks such as BiCMTS \cite{Wang2021}, which incorporate both intra- and inter-time-series attention, underscore the advantages of advanced architectures for handling missing values. Time-series models are further detailed in Section \ref{sec:related-work}. Moreover, imputation shares similarities with other data reconstruction tasks, such as rain removal from images \cite{wang2020model, long2023bishift}, highlighting the versatility of these techniques.

Self-attention mechanisms, initially developed for neural machine translation, have gained prominence in time-series imputation due to their ability to model dependencies across long temporal sequences. Recent models such as Cross-Dimensional Self-Attention (CDSA) \cite{cdsa} and DSTP-RNN \cite{Liu2020} extend these capabilities by incorporating spatial and temporal correlations effectively. These approaches demonstrate state-of-the-art (SoA) performance in various datasets by leveraging attention mechanisms for precise imputation and prediction. Several proposals \cite{saits2023, ma2019cdsa, NRTSI2023}, solely rely on self-attention networks, whereas others \cite{suo2020glima, oh2021sting}, combine self-attention with Recurrent Neural Networks (RNNs). 

In this paper, we propose \textbf{BRATI} (Bidirectional Recurrent Attention for Time-series Imputation), a novel DL model that combines bidirectional RNNs with self-attention mechanisms. BRATI addresses both short- and long-term temporal dependencies and captures feature correlations across dimensions. The contributions of this work are:
\begin{itemize}
    \item A novel combination of self-attention mechanisms and RNNs to effectively model long-term dependencies while maintaining the strengths of RNNs for short-term interactions.
    \item Extensive evaluation on three real-world datasets, covering diverse scenarios of missing data: random missing values, fixed-length sequences, and variable-length sequences.
    \item A new real-world multivariate time-series dataset, {\it Space Weather}, highlighting MNAR scenarios and serving as a benchmark for future studies.
\end{itemize}

The paper is organized as follows: Section~\ref{sec:related-work} reviews related work in time-series imputation. Section~\ref{sec:methodology} introduces the \mname\ model and its training process. Section~\ref{sec:experiments} describes the experimental setup and results. Finally, Section~\ref{sec:conclusions} presents the conclusions and future directions.

\section{Related Work}
\label{sec:related-work}

There is a considerable amount of literature in the time series imputation field. However, we only review the approaches based in DL since it has become the predominant approach. The different approaches can be categorized into 3 groups, depending on their core layers.

\begin{itemize}
    \item \textbf{RNN}: the trend to use RNNs for imputation on time series was started by Che et al. \cite{Che2018} who proposed an improvement over the Gated Recurrent Unit (GRU) called GRU-D, designed for imputing missing values in health-care data with a smooth fashion. It introduces the concept of time decay after the last valid observation, gradually diminishing the impact of the missing variable. Their proposal relies on the assumption that the missing values can be imputed combining the last observed value and the global mean. However, relying on the global mean severely limits the capabilities of the model when outliers are present. Yoon et al. proposed the M-RNN architecture \cite{mrnn} to be used in health-related datasets. It imputes the missing values using the hidden states of a bidirectional RNN. The main drawback of M-RNN is that it treats the missing values as constants, whereas another architecture, named BRITS, proposed by Cao et al. \cite{cao2018brits}, treats the missing values as variables of the graph, considering correlations among features to perform the imputation by means of bidirectional RNNs. Qian et al. \cite{qian2024deari} proposed DEARI, a Deep Attention Recurrent Neural Network that combines self-attention and Bayesian strategies to impute missing values and estimate uncertainty.

    \item \textbf{GAN}: Generative Adversarial Networks (GAN) have also been used to impute time series data. That approach usually relies on RNN, such as GRUI proposed by Luo et al. \cite{grui}. They used a modified version of the GRU Cell to handle irregular time lags and learn the distribution of incomplete time series. They also use a decay based on how much time has passed since the last observed value, similar to the one used in the GRU-D model. Later on, Luo et al. proposed E$^{2}$GAN as a follow up work \cite{luo2019e2gan}. This model features an auto-encoder, in which the generator is based on the previous GRUI to avoid generated noise. Liu et al. proposed NAOMI \cite{liu2019naomi}, a non-autoregressive model to impute long-range sequences enhanced with adversarial training, improving the imputation accuracy in trajectory prediction. Other models, such as GLIMA \cite{suo2020glima} or STING \cite{oh2021sting}, also rely on the GAN training approach with RNNs for imputation, but they are also aided by Attention Layers. Another recent example is MBGAN \cite{mbgan} which introduces the concept of temporal attention, but their proposed model only imputes one variable of the time series.
    
    \item \textbf{Self-Attention}: originally, the self-attention mechanism was proposed for sequence modeling tasks, such as neural machine translation, outperforming traditional RNN approaches. In that regard, Ma et al. \cite{ma2019cdsa} adapted the mechanism for multivariate geo-tagged time series imputation in an approach called Cross-Dimensional Self-Attention. Banshal et al. proposed DeepMVI \cite{bansal2021missing}, a self-attention based approach, aided by convolutional windows and kernel regressions to impute multivariate time-series data. Shan et al. proposed NRTSI \cite{NRTSI2023}, a method based on a Transformer encoder to impute time series sampled at irregular intervals. Du et al. proposed SAITS \cite{saits2023}, a novel method based on diagonally-masked self-attention to capture both temporal dependencies and feature correlations. They also proposed a training methodology, named Joint-optimization Training Approach to improve the traditional training procedures for imputation models. SAITS trained with their proposed approach achieved SoA results imputing values MCAR in real world datasets.
    
\end{itemize}

\section{Methodology}
\label{sec:methodology}

The proposed methodology consists of a weighted combination of two imputation blocks that process the time series in both directions. The \mname\ model is trained using the Joint-optimization training approach proposed by Du et al. \cite{saits2023} for imputing and reconstructing the time series while maintaining the consistency across both blocks of the model.

\subsection{Multivariate Time-Series with Missing values}

Given a multivariate time series with \textit{D} dimensions and \textit{T} observations, we denote it as $X = \{x_1, x_2, ..., x_T\} \in \mathbb{R}^{TxD}$. For each $t \in {1, 2, ..., T}$, $x_t \in \mathbb{R}^D$ represents the t-th observation for each of the \textit{D} dimensions, hence $x_t^d$ denotes the measurement of the \textit{d}-th variable of the \textit{t}-th time-step. The time series $X$ could have missing values, to represent them, we introduce a mask vector $M \in \mathbb{R}^{TxD}$ where,


\begin{equation}
M_t^d= \begin{cases}0 & \text { if } x_t^d \text { is missing } \\ 1 & \text{ if } x_t^d \text { is observed } \end{cases}
\label{eq:mask}
\end{equation}

\begin{figure}
\centering
\includegraphics[width=0.4\textwidth]{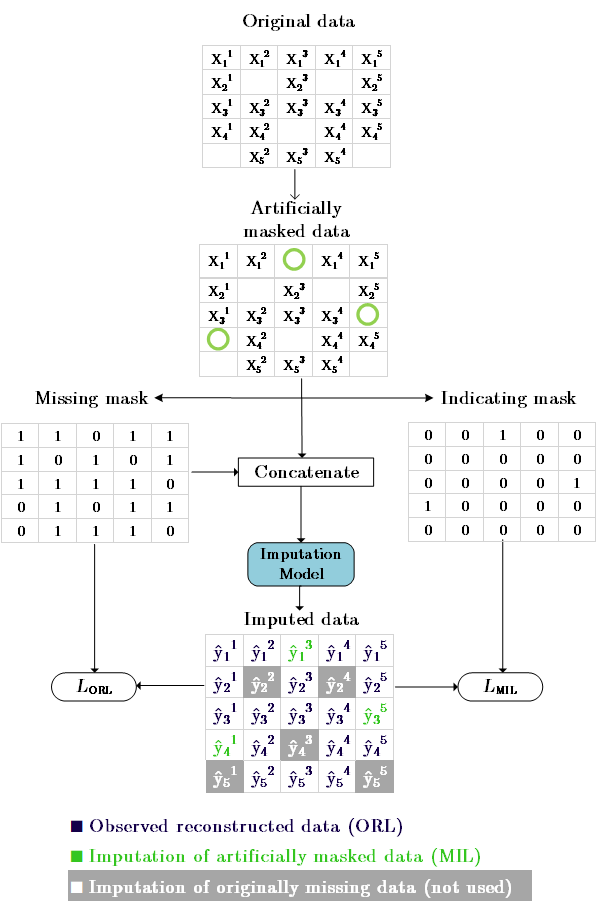}
\caption{\label{fig:masks}Graphical summary of the joint Training approach using the Masked Imputation Loss ($L_{\text{MIL}}$) and the Observed Reconstruction Loss ($L_{\text{ORL}}$).}
\end{figure}

\subsection{Joint-optimization Training approach}

We follow Du et al. \cite{saits2023} training recommendations for self-attention based imputation models. Their approach consists of two learning tasks named Masked Imputation Task (MIT) and Observed Reconstruction Task (ORT). Each task produces a loss that will be optimized during training. Those two tasks are graphically summarized in Fig. \ref{fig:masks}. In addition to those two losses, we also introduce a Consistency Task (CT), similar to the one used in the BRITS paper \cite{cao2018brits}, to force the two branches of the model to behave in a consistent manner. Each task is described as follows:

\begin{itemize}
    \item Masked Imputation Task (MIT): MIT consists of artificially masking values during training: those values will be reserved and will be fed as missing values to the model, so it will impute them. Since we have the ground truth of the missing values, the imputation loss can be calculated and the model can be optimized to impute missing values as accurately as possible. By using this approach, the number of training samples gets also naturally increased, since on each epoch the values that will be masked can be different. In order to perform this task, we mask values in the input time series obtaining $\hat{X}$ and its \textit{missing mask vector} $\hat{M}$. The output of the imputation model will be denoted as $\tilde{X}$. We also need an additional mask vector to identify those artificially masked values to calculate the imputation error later on; that vector is named \textit{indicating mask vector} and will be denoted as $I$. 
    
    After the imputation, to calculate the loss we use the Mean Absolute Error (MAE) function, but it only needs to be calculated on the artificially masked values that have been imputed, not on all the time-series. The definition of the mask vectors and the Masked Imputation Loss (MIL) calculations are depicted in Eq. \ref{eq:masks} to \ref{eq:mil-def}. The $\odot$ operation represents the element-wise product, also known as Hadamard product.
    
    \begin{equation}
        \mathbf{\hat{M}}_t^d= \begin{cases}0 & \text { if } x_t^d \text { is missing } \\ 1 & \text{ if } x_t^d \text { is observed } \end{cases} ,\text{   }
        \mathbf{I}_t^d= \begin{cases}0 & \text{ otherwise } \\ 1 & \text { if } \hat{X}_t^d \text { has been artificially masked } \end{cases}
        \label{eq:masks}
        \end{equation}
        
        \begin{equation}
        \text{MAE}_{\text{masked}} \text{(prediction, target, mask)} = \frac{\sum_d^D \sum_t^T | (prediction - target) \odot mask |_t^d}{\sum_d^D \sum_t^T mask_t^d}
        \label{eq:masked_mae}
        \end{equation}
        
        \begin{equation}
        L_{\text{MIL}} = \text{MAE}_{\text{masked}}(\tilde{X}, X, I)
        \label{eq:mil-def}
    \end{equation}
    
    \item Observed Reconstruction Task (ORT): is the traditional approach used to train imputation models for time-series. It has been widely used in the models mentioned earlier \cite{cao2018brits, luo2019e2gan, mrnn}. This approach consists of calculating the difference between the observed values fed to the model and the ones produced by the model, that is, the learned representation of the model. That representation is usually known as reconstruction. To calculate the reconstruction loss, the MAE is only calculated over the values that are not missing. The Observed Reconstruction Loss (ORL) is defined in Eq. \ref{eq:orl-def}:
    
    \begin{equation}
    L_{\text{ORL}} = \text{MAE}_{\text{masked}}(\tilde{X}, X, \hat{M})
    \label{eq:orl-def}
    \end{equation}
    
    \item Consistency Task (CT): is an approach introduced by Cao et al. when they developed BRITS \cite{cao2018brits}.
    They proposed the Consistency Loss (CL) to introduce consistency constraints so the prediction of each step would be as similar as possible in both directions. This approach helped the convergence of the network and showcased an increase in accuracy if the model followed such architecture. The CL is calculated as the MAE between the learned representation of the branch that imputes the time-series in the forward direction ($\tilde{X}_{fwd}$) and the backward one ($\tilde{X}_{bwd}$) as in Eq. \ref{eq:cl-def}.
    
    \begin{equation}
    L_{\text{CL}} = \text{MAE}(\tilde{X}_{fwd}, \tilde{X}_{bwd}).
    \label{eq:cl-def}
    \end{equation}
  
\end{itemize}

\subsection{\mname}

\begin{figure}[ht]
\centering
\includegraphics[width=0.5\textwidth]{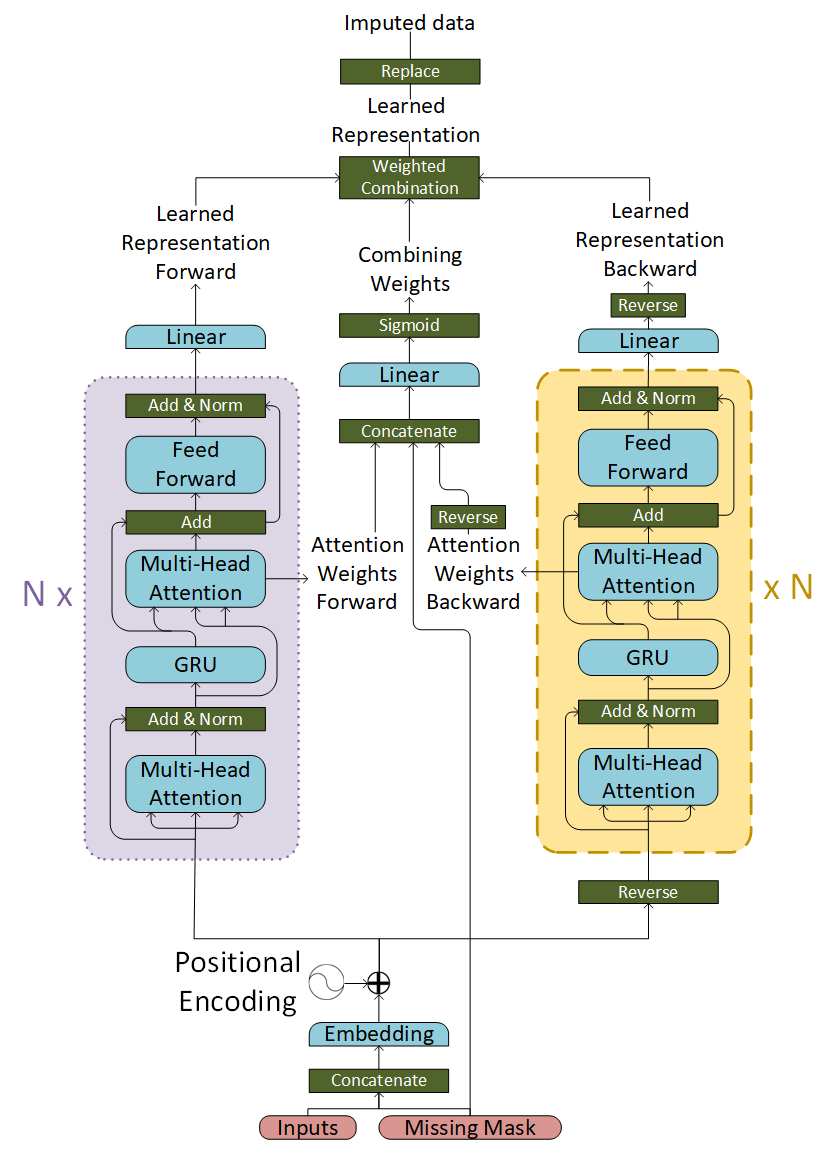}
\caption{\label{fig:arch}\mname\ Imputation architecture. The dotted purple rectangle represents the forward imputation block, the dashed yellow rectangle represents the backwards imputation blocks. The blue rounded rectangles represent operation with weights that can be optimized whereas the green straight rectangles represent operation without optimizable parameters.}
\end{figure}

Figure \ref{fig:arch} depicts the architecture of our proposed model, \mnamelong\ (\mname). It consists of two imputation blocks and a weighted combination of their representations. Each imputation block is composed of two Multi-Head Attention layers, a GRU layer and a position-wise Feed-Forward network. In both imputation blocks, the GRU layer processes the time series in different directions.

The individual components of \mname\ are explained in the subsections \ref{subsec:emb} to \ref{subsec:weighted-comb}, the compound loss function is described in the subsection \ref{subsec:loss-fn}.

\subsubsection{Embedding layer and Positional Encoder}
\label{subsec:emb}

Before the input features $\hat{X}$ are fed into the imputation blocks, they are concatenated to the missing mask vector $\hat{M}$ and projected into $d_{model}$ dimensions. Later on, we add a Positional Encoder vector, denoted as $PosEnc$ to give the Self-Attention layers information regarding the position of each value. This produces the vector $e$ that will be used by the imputation block, as noted in Eq. \ref{eq:emb}, where $e$ will be the input for the forward imputation block and $e'$ will be the input for the backward imputation block and $W_{emb} \in \mathbb{R}^{d_{\text {model }}\times D}, b_{emb} \in \mathbb{R}^{d_{\text {model }}}$. Before the vector is fed to the block that processes the series in the backwards direction, it will be reversed. To differentiate the backward block from the forward one, it will be denoted as $e'$. From now on, any mathematical symbol followed by $'$ will denote the backwards counterpart.

\begin{equation}
    \begin{aligned}
    e = [\text{Concat}(\hat{X}, \hat{M}) W_{emb} + b_{emb}] + PosEnc \\
    e' = \text{Reverse}\ e \\
    \label{eq:emb}
    \end{aligned}
\end{equation}

Following the original Transformer proposed by Vaswani et al. \cite{transformer}, we use a positional encoder to give information to the attention layers regarding the order of the individual time-steps inside the sequence. Otherwise, the Multi-Head attention layers would not make use of the order of the sequence and would treat each vale as independent of the others. In our work, we maintain the proposed fixed positional encoding based on sine and cosine functions applied at different frequencies, as defined in the Eq. \ref{eq:pos_enc}.

\begin{eqnarray}
PosEnc{(p o s, 2 i)} &=\sin \left(\frac{p o s}{10000^{\frac{2 i}{d_{\text {model }}}}}\right) \nonumber \\
PosEnc{(p o s, 2 i+1)} &=\cos \left(\frac{p o s}{10000^{\frac{2 i}{d_{\text {model }}}}}\right)
\label{eq:pos_enc}
\end{eqnarray}

\subsubsection{Attention}
\label{subsec:attn}

Similar to the original Transformer architecture proposed by Vaswani et al. \cite{transformer}, we apply the Scaled Dot-Product Attention. Despite being originally designed to tackle the machine language translation problem, it has also been successfully applied in other sequence processing tasks \cite{li2019enhancing, zhou2021informer}, or even image processing \cite{han2022survey}.

The attention function takes three vectors as an input. They are projected into a query vector, denoted as $Q$, with a dimensionality of $d_k$, and a key vector, $K$, with a dimensionality of $d_k$ and a value vector, $V$, with a dimensionality of $d_v$. Then, the attention weights are calculated using a scaled dot-product operation between $Q$ and $K$. After that, the softmax function is applied and the result is multiplied by $V$, obtaining an attention-weighted representation of $V$. The computation is described in Eq \ref{eq:att}.

\begin{equation}
    \text{Attention}(Q, K, V) = \text{softmax}(\frac{QK^T}{\sqrt{d_k}})V
    \label{eq:att}
\end{equation}

In our particular case, we will work both with self-attention, and normal attention. For the self-attention case, the vectors $Q$, $K$ and $V$ are all the same whereas on normal attention they are not all the same. Moreover, we adopt the Multi-Head approach. Instead of performing a single attention function, the vectors are projected $h$ times, where $h$ is the number of heads, and each head performs the attention function in parallel. Finally, the obtained values are concatenated and projected one final time. This approach enables the model to attend to information from different subspaces located at different positions at the same time. Multi-Head Attention is described in Eq. \ref{eq:mha}, where $head_i =$ Attention$(QW_i^Q, KW_i^K, VW_i^V)$ and the projections are parameter matrices $W_i^Q \in \mathbb{R}^{d_{\text {model }} \times d_k}, W_i^K \in \mathbb{R}^{d_{\text {model }} \times d_k}$, $W_i^V \in \mathbb{R}^{d_{\text {model }} \times d_v}$, and $W^O \in \mathbb{R}^{h d_v \times d_{\text {model }}} $.

\begin{equation}
    \begin{aligned}
    \text{MultiHeadAttention}(Q, K, V) = \text{Concat}(head_1, head_2, ..., head_h)W^O\\
    \label{eq:mha}
    \end{aligned}
\end{equation}

\subsubsection{GRU}
\label{subsec:gru}

To exploit the advantages of RNNs when working with temporal dependencies \cite{Che2018, cao2018brits}, we use a standard GRU layer, proposed by Cho et al. \cite{GRU} to process the input time-series. The input of each GRU unit is the hidden state computed by the previous unit, denoted as $h_{t-1}$, and the current input $X_t$. The GRU layer features an update gate and a reset gate that produce the output $h_t$ for the given timestep. The computation is described in Eq. \ref{eq:gru}, where $\sigma$ represents the sigmoid function, $h_t$ is the hidden state at time $t$, $x_t$ is the input at the time $t$, $h_{(t-1)}$ is the previous hidden state of the GRU (or 0 if it is the initial time-step), $r_t$, $z_t$ and $n_t$ are the reset, update and new gates, respectively. 

\begin{eqnarray}
    r_t = \sigma(W_{i r} x_t+b_{i r}+W_{h r} h_{(t-1)}+b_{h r}\nonumber\\
    z_t =\sigma(W_{i z} x_t+b_{i z}+W_{h z} h_{(t-1)}+b_{h z}\nonumber\\
    n_t =\tanh (W_{i n} x_t+b_{i n}+r_t \odot (W_{h n} h_{(t-1)}+b_{h n})) \nonumber\\
    h_t = (1-z_t) \odot n_t+z_t \odot h_{(t-1)} 
    \label{eq:gru}
\end{eqnarray}


\subsubsection{Feed-Forward network}

In addition to the GRU and Attention Layers, we complement them with a position-wise Feed-Forward network, as is a common practice in any Transformer-based model. It consists of two Linear layers, the first one with a different dimensionality than the input and the second one with the same dimensionality as the input to the Feed-Forward Network, so we can employ residual connections. Particularly, we use the ELU activation function between the linear layers as shown in Eq. \ref{eq:ffn}. We chose the ELU activation function instead of the more usual ReLU function since it showed a slight increase in performance. This operation is applied to each time-step separately.

\begin{eqnarray}
    \text{FFN}(x) = \text{ELU}(xW_1 + b_1) W_2 + b_2 \label{eq:ffn}\\
    \text{where } W_1 \in \mathbb{R}^{d_{\text {model }} \times d_{\text {ffn }}}, b_1 \in \mathbb{R}^{d_{\text {ffn }}}, W_2 \in \mathbb{R}^{d_{\text {ffin }} \times d_{\text {model }}}, b_2 \in \mathbb{R}^{d_{\text {model }}}\nonumber
\end{eqnarray}

\subsubsection{Imputation blocks}

The internal architecture of both Imputation Blocks is the same, the only difference is that the forward Imputation Block takes $e$ as the input, while the Backwards one takes $e'$, both vectors obtained as in Eq. \ref{eq:emb}.

First, the embedded vector $e$ is reweighted as in Eq. \ref{eq:ibmha} using a Multi-Head Attention layer, following Eq. \ref{eq:mha}, to produce $\alpha$. In this case, the Attention layer will work as a Self-Attention layer, since all the input vectors, Q, K and V are the same, $e$. Then, $\alpha$ is recursively processed by the GRU layer as in Eq. \ref{eq:ibgru}, following Eq. \ref{eq:gru}, to produce $\beta$. The next Multi-Head Attention Layer will perform the Attention operation over both $\alpha$ and $\beta$, using $\alpha$ as the Query Vector ($Q$) and $\beta$ as the Key and Value vectors ($K$, $V$), following Eq. \ref{eq:ibcmha}. This equation produces $\gamma$, which is the output of the Multi-Head Attention layer, and $\gamma_w$, which denotes the attention weights averaged across all the heads. The attention weights will be used later in the weighted combination of both Imputation Blocks. On the last step, the Feed-Forward network will produce $\delta$ using the addition of $\gamma$ and $\beta$ as input as in Eq \ref{eq:ibffn}\footnote{Layer normalization \cite{layer_norm} and Residual connection \cite{res_connection} operations are omitted here for simplicity, but they are included in Fig. \ref{fig:arch}.}. The Imputation Blocks can be stacked $N$ times. Each produced $\delta$ will be the input for the following block ($e$). The last $\delta$ and $\gamma_w$ will be the ones used in the next section.

\begin{eqnarray}
\alpha = \text{MultiHeadAttention}(e, e, e) \label{eq:ibmha}\\
\beta = \text{GRU}(\alpha) \label{eq:ibgru}\\
\gamma, \gamma_{w} = \text{MultiHeadAttention}(\beta, \alpha, \alpha) \label{eq:ibcmha}\\
\delta = \text{FFN}(\gamma + \beta) \label{eq:ibffn}
\end{eqnarray}

\subsubsection{Weighted Combination}
\label{subsec:weighted-comb}

After each Imputation block, we will reduce the dimensionality of the output of the Imputation from $d_\text{model}$ to $D$, the original dimensionality of $X$, where $W_z \in \mathbb{R}^{d_{model}\times D}$, $b_z \in \mathbb{R}^D$. This produces the learned representation of the input vector, $\hat{X}$, as in Eq \ref{eq:reduce_dim}, since each Imputation block produces a different representation. In Eq. \ref{eq:weighted_comb} we use the attention weights produced by each block $\gamma_{w}$ and $\gamma'_{w}$, together with the input mask $\hat{M}$ to produce the combining weights $\Omega \in (0, 1)^{T \times D}$. Where $W_{\Omega} \in \mathbb{R}^{(T \cdot 2 + D) \times D} \text{ and } b_{\Omega} \in \mathbb{R}^D$ are learnable parameters. Then, in Eq. \ref{eq:comb} the representations of each block, $\hat{X}$ and $\hat{X'}$, will be combined into the joint-representation $\hat{X}_j$ using the combining weights from Eq. \ref{eq:comb}. To obtain the final imputation $\hat{I}$, the missing mask $M$ is used to substitute the missing values in the original input vector $X$ for the imputed values obtained by the model as in Eq. \ref{eq:imp}.

\begin{eqnarray}
    \hat{X} = (W_z \delta ) + b_z \label{eq:reduce_dim}\\
    \Omega = \text{Sigmoid} ( W_{\Omega} \cdot ( M | \gamma_{w} | \gamma'_{w} ) + b_{\Omega} )    \label{eq:weighted_comb}\\
    \hat{X}_j = \hat{X} \odot \Omega + \hat{X'} \odot (1 - \Omega)   \label{eq:comb}\\
    \hat{I} = \hat{X} \odot \hat{M} + \hat{X}_j \odot (1 - \hat{M})   \label{eq:imp}
\end{eqnarray}

\subsubsection{Loss functions}
\label{subsec:loss-fn}

The loss that our model tries to minimize is composed of three different losses, the Masked Imputation Loss (MIL), depicted in Eq. \ref{eq:mil}, the Observed Reconstruction Loss (ORL), in Eq. \ref{eq:orl} and a Consistency Loss, in Eq. \ref{eq:consistency}. The MIL and ORL follow the joint-optimization training approach proposed by Du et al. described earlier. Additionally, we also make use of the Consistency Loss, multiplied by a consistency factor $\rho$, introduced by Cao et al. \cite{cao2018brits}, to try to make the model as consistent as possible in both directions and to help it to converge faster. Finally, we combine all the tree losses, each one multiplied by a weight which can be tuned to achieve a better trade-off between the loss functions, represented as $\lambda$: $L_{\text{MIL}} \cdot \lambda_{\text{MIL}}$, $L_{\text{ORL}} \cdot \lambda_{\text{ORL}}$ and $L_{\text{cons}} \cdot \lambda_{\text{cons}}$, into the total loss $L_{\text{total}}$, which will be minimized to improve the model's performance, as in Eq. \ref{eq:total}.

\begin{eqnarray}
    L_{\text{MIL}} = \text{MAE}_{\text{masked}}(\hat{X}_j, X, I) 
    \label{eq:mil}\\
    L_{\text{ORL}} = \frac{1}{2} \cdot \text{MAE}_{\text{masked}}(\hat{X}_j, X, M)  + \frac{1}{4} \cdot \text{MAE}_{\text{masked}} (\hat{X}, X, M) + \frac{1}{4} \cdot \text{MAE}_{\text{masked}}(\hat{X'}, X, M) 
    \label{eq:orl}\\
    L_{\text{cons}} = \text{Discrepancy}(\hat{X}, \hat{X'}) \cdot \rho
    \label{eq:consistency}\\
    L_{\text{total}} = L_{\text{MIL}} \cdot \lambda_{\text{MIL}} + L_{\text{ORL}} \cdot \lambda_{\text{ORL}} + L_{\text{cons}} \cdot \lambda_{\text{cons}}
    \label{eq:total}
\end{eqnarray}

\section{Experiments}
\label{sec:experiments}

This section describes the experimental evaluation to compare the performance of our proposed architecture to previous existing models. First, the different ways in which the artificial missing values are generated, are explained. Next, the three used datasets are presented, indicating the division in the training, validation, and testing subsets. After that, the baseline models with which we compare our model are presented. Then, we introduce the evaluation metrics and the training settings for each model. Finally, the experimental results are presented.

\subsection{Artificial generation of missing values}
\label{subsec:generation-missing}

Since the real values of the missing data are not available to evaluate the performance of the imputation models, we artificially remove parts of the observed values in the validation and test sets and use these values as ground truth for evaluation. The goal is to train the models to handle both independent missing values and sequences of missing values, as sequences are commonly found in the target datasets. The values removed and used as ground truth are selected to simulate different missing patterns. To ensure a comprehensive evaluation, 10\% and 20\% of the observed data are removed for each pattern, simulating the following scenarios:

\begin{enumerate}
    \item \textbf{Independent missing values}: values are randomly removed using a uniform distribution until 10\% or 20\% of the observed values are eliminated, simulating a MCAR scenario.
    \item \textbf{Missing in a sequence of fixed length}: instead of randomly removing independent values, sequences of 5 consecutive observed values are removed. If there are insufficient sequences of 5 consecutive observed values, additional values are randomly removed to meet the target elimination rate of 10\% or 20\%.
    \item \textbf{Missing in a sequence of random length}: sequences of observed values with lengths randomly varying between 3 and 10 are removed. The length of each sequence is chosen from a uniform distribution. If there are not enough sequences of consecutive observed values, additional values are randomly removed as in the first scenario.
\end{enumerate}

Experimental results show that the second and third scenarios are considerably more difficult across all the evaluation metrics than the MCAR one, being the last case the most difficult one.

\subsection{Datasets}
\label{subsec:datasets}

We evaluate our proposed model and the baseline methods on three different real-world datasets from different domains: the PhysioNet 2012 Mortality Prediction Challenge dataset\footnote{\url{https://www.physionet.org/content/challenge-2012}}, a {\it Water Quality} dataset\footnote{\url{https://github.com/niqingjian2021/Water-Quality-Data}}, and a Space Weather measurements dataset. General information regarding these datasets is summarized in Table \ref{tab:gen-info-datasets}. All the data have been standardized during the preprocessing.

\begin{table}[]
\centering
\caption{General information of the evaluated datasets}
\label{tab:gen-info-datasets}
\begin{tabular}{|c|c|c|c|}
\hline
                      & PhysioNet-2012 & Water Quality & Space Weather \\ \hline
Samples               & 11,988         & 27,203        & 129,168       \\ \hline
Features              & 37             & 5             & 7             \\ \hline
Sequence length       & 48             & 24            & 24            \\ \hline
Original missing rate & 80.0\%         & 1.1\%         & 4.4\%         \\ \hline
\end{tabular}
\end{table}

\begin{enumerate}
    \item \textbf{PhysioNet:} this dataset contains healthcare data from the PhysioNet Challenge 2012~\cite{PhysioNet, physionet2}. It consists of 12,000 records from de-identified patients in Intensive Care Units (ICU), covering approximately 48 hours of multivariate time-series data with 37 features, including blood pressure, temperature, and respiration rate. Measurements are recorded irregularly, with some collected at regular intervals (hourly or daily) and others, such as specific tests, only on-demand. This dataset has been widely used in time-series imputation works, such as SAITS \cite{saits2023} and BRITS \cite{cao2018brits}.
    
    For evaluation, the dataset is divided into 80\% for training/validation and 20\% for testing. In the validation and test sets, 10\% or 20\% of the observed data is artificially removed in the three ways mentioned in Section \ref{subsec:generation-missing}, resulting in 6 configurations.

    \item \textbf{Water Quality:} this dataset comprises measurements of 5 water quality attributes (water temperature, dissolved oxygen, conductivity, turbidity, and pH) from a monitoring station in Jiangsu province, China. The data spans from August 2017 to October 2021, recorded hourly, with some missing values in all attributes. This dataset is referenced in the MBGAN paper~\cite{mbgan}.
    
    The dataset is split similarly to {\it PhysioNet}, with 80\% for training/validation and 20\% for testing. In the validation and test sets, 10\% or 20\% of the observed data is removed artificially, following the patterns described in Section \ref{subsec:generation-missing}.

    \item \textbf{Space Weather:} this dataset includes measurements of the Interplanetary Magnetic Field (IMF) and solar wind plasma collected by the ACE space probe \cite{stone1998advanced}. Data is available since 1998, but most measurements are stable except during geomagnetic storms, which disturb Earth's magnetosphere and disrupt systems \cite{morina2019probability}. During storms, plasma instruments often saturate, leaving measurements unavailable. However, IMF measurements remain almost fully available, providing related information for imputation.

    The dataset focuses on geomagnetic storms when plasma values are most frequently missing, rather than using the complete twenty-year solar wind dataset. Following similar approaches to previous forecasting studies \cite{Siciliano2021, collado2024operational}, we split the storms into 20 for training, 5 for validation, and 17 for testing. The plasma data is obtained from the \texttt{AC\_H0\_SWE} dataset, while the IMF data comes from the \texttt{AC\_H3\_MFI}, both accessible via NASA's CDAWeb ~\footnote{\url{https://cdaweb.gsfc.nasa.gov/}}.  
\end{enumerate}

\subsection{Baseline methods}

To evaluate the performance of our proposed model, we compare it against two naive imputation methods and four DL models that have previously achieved SoA results. Although the {\it Water Quality} dataset has been used to evaluate the MBGAN architecture~\cite{mbgan}, we chose not to include MBGAN in the comparison because it is limited to imputing a single variable, whereas the three datasets under evaluation present missing values in multiple variables.

\begin{itemize}
    \item \textbf{Naive imputation methods:} we employ two basic strategies for comparison: 
    \begin{itemize}
        \item \textit{Median imputation:} missing values are filled using the median value of each column, calculated over the training set. 
        \item \textit{Last observation carried forward:} missing values are filled with the last observed value for each column. If there are no previous observations, the missing value is filled with 0.
    \end{itemize}
    
    \item \textbf{Deep Learning models:} we compare our proposed model to the following:
    \begin{itemize}
        \item \textit{M-RNN:} a model based on bidirectional RNNs designed for multivariate time-series imputation \cite{mrnn}.
        \item \textit{BRITS:} a method that models missing values as variables in a bidirectional RNN framework, considering feature correlations \cite{cao2018brits}.
        \item \textit{Transformer encoder model:} a model based on the Transformer architecture for capturing long-term dependencies \cite{transformer}.
        \item \textit{SAITS:} A diagonally-masked self-attention-based model for multivariate time-series imputation \cite{saits2023}.
    \end{itemize}

\end{itemize}

\subsection{Experimental settings}

We use three metrics to evaluate the imputation performance of the models: Mean Absolute Error (MAE), Root Mean Squared Error (RMSE), and Mean Relative Error (MRE). These metrics are only calculated for the values that have been artificially masked in the test sets, as the ground truth is known for these values. This corresponds to the positions where the Indicating Mask $I$, introduced earlier, is equal to 1. The metrics are defined in Eq. \ref{eq:mae} to \ref{eq:mre}.

\begin{equation}
    \text{MAE}(\text{imputation}, \text{target}, \text{mask}) = \frac{\sum^D_d \sum^T_t |(\text{imputation} - \text{target}) \odot \text{mask}|^d_t}{\sum^D_d \sum^T_t \text{mask}^d_t}  \\
    \label{eq:mae}
\end{equation}

\begin{equation}
    \text{RMSE}(\text{imputation}, \text{target}, \text{mask}) = \sqrt{\frac{\sum^D_d \sum^T_t (((\text{imputation} - \text{target}) \odot \text{mask})^2)^d_t}{\sum^D_d \sum^T_t \text{mask}^d_t}} \\ 
    \label{eq:rmse}
\end{equation}

\begin{equation}
    \text{MRE}(\text{imputation}, \text{target}, \text{mask}) = \frac{\sum^D_d \sum^T_t |(\text{imputation} - \text{target}) \odot \text{mask}|^d_t}{\sum^D_d \sum^T_t |\text{target} \odot \text{mask}|^d_t}
    \label{eq:mre}
\end{equation}

To ensure a fair comparison, all models are trained using the Adam optimizer with $\beta_1 = 0.9$, $\beta_2 = 0.98$, and $\epsilon = 10^{-9}$, following the configuration proposed by Vaswani et al. \cite{transformer}. The learning rate is scheduled using Eq. \ref{eq:scheduler}, with \texttt{warmup\_steps} set to 4000. This scheduler eliminates the need to search for an additional learning rate hyperparameter, as it adjusts the learning rate based on the model dimensionality. Training is stopped early if the imputation MAE does not decrease after 30 epochs, and the best validation weights are restored before evaluating the test dataset.

\begin{equation}
    lr = d^{-0.5}_{\text{model}} \cdot \text{min}(\text{step\_num}^{-0.5},\ \text{step\_num} \cdot \text{warmup\_steps}^{-1.5})
    \label{eq:scheduler}
\end{equation}

Moreover, a hyperparameter search is performed for each model and dataset using Microsoft's Neural Network Intelligence system \cite{nni}, which implements the Tree-structured Parzen Estimator (TPE), a Sequential Model-Based Optimization (SMBO) algorithm. TPE models the conditional probability $P(x|y)$ and the marginal probability $P(y)$, where $x$ represents the hyperparameters and $y$ the evaluation result. It substitutes the configuration priors with non-parametric densities for efficient optimization \cite{bergstra2011algorithms}.

The details of the hyperparameter search are defined as follows:

\begin{itemize}
    \item \textbf{RNN-based models (MRNN, BRITS):} the RNN hidden size is sampled from (32, 64, 128, 256, 512, 1024).
    \item \textbf{Self-Attention models (Transformer, SAITS):} the number of encoder blocks, $N$ is sampled from (1, 2, 4, 5, 6), the dimensionality of the model $d_{model}$ is sampled from (64, 128, 256, 512), the dimensionality of the feed-forward network, $d_{\text{ffn}}$ is sampled from (128, 256, 512, 1024), the dimensionality of the key and value vectors $d_k$ and $d_v$ for the attention layers are sampled from (32, 64, 128, 256), the number of heads is sampled from (2,4,8). The dropout is sampled from (0, 0.1, 0.2, 0.3, 0.4, 0.5). The dropout is sampled from (0, 0.1, 0.2, 0.3, 0.4, 0.5).
    \item \textbf{\mname\ model:} the number of Imputation blocks $N$ are sampled from (1, 2, 3, 4), the dimensionality of the model $d_{model}$ is sampled from (64, 128, 256, 512), the dimensionality of the key and value vectors, namely $d_k$ and $d_v$ in the other Self-Attention models is set as the value of $d_{model}$, the dimensionality of the feed-forward network, $d_{\text{ffn}}$, is sampled from (128, 256, 512, 1024) the number of heads is sampled from (2,4,8). The dropout is sampled from (0, 0.1, 0.2, 0.3, 0.4, 0.5), we also set as a hyperparameter whether to add a bias vector in the GRU and the Multi-Head Attention layers. The consistency factor $\rho$ used to weigh the consistency loss has been set to 0.1.
\end{itemize}

\subsection{Experimental results}

The main limitation of RNNs lies in their tendency to treat all input elements equally, which can lead to suboptimal performance when sequences are very long or when certain time-steps carry more informative content. Attention layers address this limitation by dynamically reweighting the importance of each input element before processing the sequence. By combining RNNs with Attention layers, models can better capture context and dependencies, improving their ability to impute missing values effectively.

We use the 3 datasets described in Section \ref{subsec:datasets} to evaluate and compare the imputation performance of \mname. The results are summarized in Tables \ref{tab:physionet_part1} and \ref{tab:physionet_part2} for the {\it PhysioNet} dataset, Tables \ref{tab:water_part1} and \ref{tab:water_part2} for the {\it Water Quality} dataset, and Tables \ref{tab:space_weather_part1} and \ref{tab:space_weather_part2} for the {\it Space Weather} dataset. Below, we provide a detailed analysis of the results for each dataset.

\paragraph{PhysioNet dataset}  
Tables  \ref{tab:physionet_part1} and \ref{tab:physionet_part2} report the metrics for the {\it PhysioNet} dataset, the most challenging due to its high sparsity. Imputation becomes significantly harder across all models when missing values appear in sequences, as these configurations disrupt temporal dependencies. MRNN outperforms Median imputation but is surpassed by propagating the last observed value. BRITS, leveraging its recurrent structure, captures temporal dependencies more effectively, particularly in scenarios with sequences of fixed length, and outperforms both Naive imputation methods.

Self-Attention models (Transformer and SAITS) demonstrate superior performance compared to RNN-based methods. While SAITS significantly outperforms the Transformer model in the MCAR scenario, their performance is comparable in other configurations. \mname\ consistently achieves the best results across all configurations, with the largest improvements observed in scenarios with missing values in sequences of random length, highlighting its ability to handle complex missing patterns effectively.

\paragraph{Water Quality dataset}  
Tables \ref{tab:water_part1} and \ref{tab:water_part2} show the results for the {\it Water Quality} dataset. For the RNN-based models, the results follow a similar trend to the {\it PhysioNet} dataset: MRNN outperforms Median imputation, and BRITS achieves better performance than both Naive methods in most configurations. However, propagating the last observed value yields better results for the most challenging scenario, missing values in sequences of random length.

The Self-Attention models exhibit peculiar behavior, struggling with convergence in the MCAR scenario, where BRITS outperforms them. This suggests that RNN-based models, including \mname, handle this configuration more effectively. However, in scenarios with missing values in sequences of fixed and random lengths, the Self-Attention models significantly outperform RNN-based methods, delivering performance very close to \mname. Even so, \mname\ achieves the best results across all six configurations, showing notable improvements in MAE and MRE metrics for the MCAR and fixed sequence length cases. For sequences of random length, while the differences in MAE and MRE metrics between \mname\ and the Self-Attention models are negligible, \mname\ demonstrates a clear advantage in the RMSE metric, particularly when 10\% of the data is artificially masked.

\paragraph{Space Weather dataset}  
Tables \ref{tab:space_weather_part1} and \ref{tab:space_weather_part2} present the evaluation metrics for the {\it Space Weather} dataset. Similar to previous datasets, MRNN outperforms Median imputation on the RMSE metric in the MCAR scenario and significantly improves in the fixed sequence length case. However, its performance declines for missing values in sequences of random length. BRITS achieves substantial improvements over the Naive methods across all configurations.

Self-Attention models demonstrate a significant advantage over RNN-based methods, with SAITS outperforming the Transformer model in all scenarios. \mname\ achieves the best performance across all metrics and configurations, with the largest improvements observed for scenarios with missing values in sequences of random length. The recurrent nature of the GRU layer in \mname's imputation blocks proves particularly effective in handling these challenging configurations.

\begin{table}[!h]
    \centering
    \caption{Performance comparison between the baseline methods and \mname\ on the {\it PhysioNet} dataset (Part 1).\label{tab:physionet_part1}}
    \renewcommand{\arraystretch}{1.2}
    \begin{tabular}{|l|P{32pt}|P{32pt}|P{32pt}|P{32pt}|P{32pt}|P{32pt}|P{32pt}|P{32pt}|P{32pt}|}
    \hline
         \textbf{PhysioNet} & \multicolumn{3}{c|}{0.1 random} & \multicolumn{3}{c|}{0.2 random} & \multicolumn{3}{c|}{0.1 series 5} \\ \hline
        \textit{Metric} & MAE & RMSE & MRE & MAE & RMSE & MRE & MAE & RMSE & MRE \\ \hline
        Median & 0.6823 & 1.0077 & 0.9790 & 0.6835 & 0.9885 & 0.9777 & 0.7095 & 0.9639 & 0.9801 \\ \hline
        Last value & 0.4083 & 0.7676 & 0.5859 & 0.4177 & 0.7642 & 0.5974 & 0.4903 & 0.8033 & 0.6772 \\ \hline
        MRNN & 0.5277 & 0.8061 & 0.7572 & 0.5436 & 0.7951 & 0.7775 & 0.5747 & 0.7820 & 0.7938 \\ \hline
        BRITS & 0.2333 & 0.6070 & 0.3448 & 0.2534 & 0.5729 & 0.3625 & 0.3271 & 0.5285 & 0.4518 \\ \hline
        Transformer & 0.1935 & 0.4846 & 0.2776 & 0.2113 & 0.4609 & 0.3023 & 0.2535 & 0.4499 & 0.3501 \\ \hline
        SAITS & 0.1893 & 0.4846 & 0.2716 & 0.1997 & 0.4422 & 0.2857 & 0.2522 & 0.4517 & 0.3484 \\ \hline
        \mname\ & \textbf{0.1854} & \textbf{0.4786} & \textbf{0.2660} & \textbf{0.1974} & \textbf{0.4404} & \textbf{0.2823} & \textbf{0.2507} & \textbf{0.4493} & \textbf{0.3463} \\ \hline
    \end{tabular}
\end{table}

\begin{table}[!h]
    \centering
    \caption{Performance comparison between the baseline methods and \mname\ on the {\it PhysioNet} dataset (Part 2).\label{tab:physionet_part2}}
    \renewcommand{\arraystretch}{1.2}
    \begin{tabular}{|l|P{32pt}|P{32pt}|P{32pt}|P{32pt}|P{32pt}|P{32pt}|P{32pt}|P{32pt}|P{32pt}|}
    \hline
         \textbf{PhysioNet} & \multicolumn{3}{c|}{0.2 series 5} & \multicolumn{3}{c|}{0.1 range 3-10} & \multicolumn{3}{c|}{0.2 range 3-10} \\ \hline
        \textit{Metric} & MAE & RMSE & MRE & MAE & RMSE & MRE & MAE & RMSE & MRE \\ \hline
        Median & 0.7032 & 0.9676 & 0.9795 & 0.6902 & 0.9820 & 0.9775 & 0.6830 & 0.9791 & 0.9767 \\ \hline
        Last value & 0.5107 & 0.8416 & 0.7114 & 0.5663 & 0.9230 & 0.8020 & 0.6357 & 1.0402 & 0.9090 \\ \hline
        MRNN & 0.6080 & 0.8282 & 0.8468 & 0.6274 & 0.8754 & 0.8884 & 0.6468 & 0.8992 & 0.9249 \\ \hline
        BRITS & 0.3687 & 0.5862 & 0.5135 & 0.4383 & 0.6911 & 0.6207 & 0.5089 & 0.7664 & 0.7277 \\ \hline
        Transformer & 0.2962 & 0.5097 & 0.4126 & 0.3534 & 0.6025 & 0.5005 & 0.4358 & 0.7011 & 0.6232 \\ \hline
        SAITS & 0.2965 & 0.5116 & 0.4130 & 0.3556 & 0.6072 & 0.5036 & 0.4321 & 0.6998 & 0.6179 \\ \hline
        \mname\ & \textbf{0.2925} & \textbf{0.5096} & \textbf{0.4074} & \textbf{0.3463} & \textbf{0.5999} & \textbf{0.4904} & \textbf{0.4211} & \textbf{0.6945} & \textbf{0.6021} \\ \hline
    \end{tabular}
\end{table}

\begin{table}[!h]
    \centering
    \caption{Performance comparison between the baseline methods and \mname\ on the {\it Water Quality} dataset (Part 1).\label{tab:water_part1}}
    \renewcommand{\arraystretch}{1.2}        
    \begin{tabular}{|l|P{28pt}|P{28pt}|P{28pt}|P{28pt}|P{28pt}|P{28pt}|P{28pt}|P{28pt}|P{28pt}|}
    \hline
         \textbf{Water Quality} & \multicolumn{3}{c|}{0.1 random} & \multicolumn{3}{c|}{0.2 random} & \multicolumn{3}{c|}{0.1 series 5} \\ \hline
        \textit{Metric} & MAE & RMSE & MRE & MAE & RMSE & MRE & MAE & RMSE & MRE \\ \hline
        Median & 0.4730 & 0.6735 & 0.9365 & 0.4737 & 0.6743 & 0.9373 & 0.4767 & 0.6776 & 0.9373 \\ \hline
        Last value & 0.0521 & 0.1921 & 0.1032 & 0.0561 & 0.2035 & 0.1110 & 0.0792 & 0.2284 & 0.1556 \\ \hline
        MRNN & 0.3001 & 0.4216 & 0.5941 & 0.3212 & 0.4481 & 0.6356 & 0.3031 & 0.4240 & 0.5959 \\ \hline
        BRITS & 0.0435 & 0.1238 & 0.0862 & 0.0505 & 0.1347 & 0.0999 & 0.0687 & 0.1490 & 0.1351 \\ \hline
        Transformer & 0.0565 & 1.3863 & 0.1118 & 0.0599 & 1.2574 & 0.1185 & 0.0311 & 0.1514 & 0.0612 \\ \hline
        SAITS & 0.0607 & 1.4268 & 0.1203 & 0.0594 & 1.4463 & 0.1175 & 0.0321 & 0.2280 & 0.0631 \\ \hline
        \mname\ & \textbf{0.0206} & \textbf{0.1046} & \textbf{0.0408} & \textbf{0.0216} & \textbf{0.1029} & \textbf{0.0427} & \textbf{0.0275} & \textbf{0.1004} & \textbf{0.0540} \\ \hline
    \end{tabular}    
\end{table}

\begin{table}[!h]
    \centering
    \caption{Performance comparison between the baseline methods and \mname\ on the {\it Water Quality} dataset (Part 2).\label{tab:water_part2}}
    \renewcommand{\arraystretch}{1.2}        
    \begin{tabular}{|l|P{28pt}|P{28pt}|P{28pt}|P{28pt}|P{28pt}|P{28pt}|P{28pt}|P{28pt}|P{28pt}|}
    \hline
         \textbf{Water Quality} & \multicolumn{3}{c|}{0.2 series 5} & \multicolumn{3}{c|}{0.1 range 3-10} & \multicolumn{3}{c|}{0.2 range 3-10} \\ \hline
        \textit{Metric} & MAE & RMSE & MRE & MAE & RMSE & MRE & MAE & RMSE & MRE \\ \hline
        Median & 0.4740 & 0.6736 & 0.9376 & 0.4735 & 0.6736 & 0.9378 & 0.4730 & 0.6732 & 0.9373 \\ \hline
        Last value & 0.0792 & 0.2338 & 0.1567 & 0.1528 & 0.3505 & 0.3027 & 0.2569 & 0.4788 & 0.5091 \\ \hline
        MRNN & 0.3301 & 0.4578 & 0.6529 & 0.5669 & 0.7512 & 1.1230 & 0.6042 & 0.8059 & 1.1971 \\ \hline
        BRITS & 0.0751 & 0.1511 & 0.1485 & 0.1923 & 0.3000 & 0.3809 & 0.2967 & 0.4434 & 0.5879 \\ \hline
        Transformer & 0.0319 & 0.2444 & 0.0630 & 0.0509 & 0.2939 & 0.1008 & 0.0660 & 0.3237 & 0.1308 \\ \hline
        SAITS & 0.0439 & 0.6124 & 0.0869 & 0.0462 & 0.1893 & 0.0916 & 0.0659 & 0.1647 & 0.1305 \\ \hline
        \mname\ & \textbf{0.0278} & \textbf{0.0920} & \textbf{0.0551} & \textbf{0.0460} & \textbf{0.1385} & \textbf{0.0911} & \textbf{0.0629} & \textbf{0.1536} & \textbf{0.1246} \\ \hline
    \end{tabular}    
\end{table}

\begin{table}[!h]
    \centering
    \caption{Performance comparison between the baseline methods and \mname\ on the {\it Space Weather} dataset (Part 1).\label{tab:space_weather_part1}}
    \renewcommand{\arraystretch}{1.2}    
    \begin{tabular}{|l|P{28pt}|P{28pt}|P{28pt}|P{28pt}|P{28pt}|P{28pt}|P{28pt}|P{28pt}|P{28pt}|}
    \hline
         \textbf{Space Weather} & \multicolumn{3}{c|}{0.1 random} & \multicolumn{3}{c|}{0.2 random} & \multicolumn{3}{c|}{0.1 series 5} \\ \hline
        \textit{Metric} & MAE & RMSE & MRE & MAE & RMSE & MRE & MAE & RMSE & MRE \\ \hline
        Median & 0.6044 & 0.9010 & 0.9597 & 0.6048 & 0.9018 & 0.9597 & 0.6025 & 0.8975 & 0.9587 \\ \hline
        Last value & 0.1226 & 0.2850 & 0.1946 & 0.1294 & 0.2984 & 0.2054 & 0.1839 & 0.3989 & 0.2926 \\ \hline
        MRNN & 0.6180 & 0.8958 & 0.9813 & 0.6220 & 0.9003 & 0.9870 & 0.6166 & 0.8920 & 0.9812 \\ \hline
        BRITS & 0.1024 & 0.2113 & 0.1626 & 0.1123 & 0.2252 & 0.1782 & 0.1606 & 0.2961 & 0.2556 \\ \hline
        Transformer & 0.0606 & 0.1413 & 0.0962 & 0.0606 & 0.1440 & 0.0961 & 0.0744 & 0.1998 & 0.1183 \\ \hline
        SAITS & 0.0501 & 0.1274 & 0.0795 & 0.0542 & 0.1363 & 0.0860 & 0.0718 & 0.1962 & 0.1142 \\ \hline
        \mname\ & \textbf{0.0471} & \textbf{0.1255} & \textbf{0.074}8 & \textbf{0.0528} & \textbf{0.1354} & \textbf{0.0838} & \textbf{0.0680} & \textbf{0.1881} & \textbf{0.1083} \\ \hline
    \end{tabular}
\end{table}

\begin{table}[!h]
    \centering
    \caption{Performance comparison between the baseline methods and \mname\ on the {\it Space Weather} dataset (Part 2).\label{tab:space_weather_part2}}
    \renewcommand{\arraystretch}{1.2}    
    \begin{tabular}{|l|P{28pt}|P{28pt}|P{28pt}|P{28pt}|P{28pt}|P{28pt}|P{28pt}|P{28pt}|P{28pt}|}
    \hline
         \textbf{Space Weather} & \multicolumn{3}{c|}{0.2 series 5} & \multicolumn{3}{c|}{0.1 range 3-10} & \multicolumn{3}{c|}{0.2 range 3-10} \\ \hline
        \textit{Metric} & MAE & RMSE & MRE & MAE & RMSE & MRE & MAE & RMSE & MRE \\ \hline
        Median & 0.6013 & 0.8927 & 0.9572 & 0.6025 & 0.9003 & 0.9586 & 0.6031 & 0.8962 & 0.9585 \\ \hline
        Last value & 0.1863 & 0.3959 & 0.2965 & 0.2097 & 0.4410 & 0.3336 & 0.2154 & 0.4449 & 0.3433 \\ \hline
        MRNN & 0.6235 & 0.8969 & 0.9926 & 0.6205 & 0.9004 & 0.9873 & 0.6299 & 0.9059 & 1.0010 \\ \hline
        BRITS & 0.1700 & 0.2981 & 0.2707 & 0.1882 & 0.3308 & 0.2995 & 0.2061 & 0.3449 & 0.3275 \\ \hline
        Transformer & 0.0835 & 0.2070 & 0.1329 & 0.0889 & 0.2229 & 0.1415 & 0.0974 & 0.2286 & 0.1547 \\ \hline
        SAITS & 0.0798 & 0.2018 & 0.1271 & 0.0824 & 0.2159 & 0.1311 & 0.0953 & 0.2317 & 0.1514 \\ \hline
        \mname\ & \textbf{0.0754} & \textbf{0.1931} & \textbf{0.1200} & \textbf{0.0787} & \textbf{0.2118} & \textbf{0.1252} & \textbf{0.0932} & \textbf{0.2291} & \textbf{0.1482} \\ \hline
    \end{tabular}
\end{table}

\subsection{Ablation experiment}

In this section, we conduct two experiments to analyze the impact of key components in the \mname\ architecture. The first experiment evaluates the performance of the model when only one imputation block is used, while the second experiment focuses on the use of combining weights derived from attention. These experiments are performed on all three datasets across their six configurations. The hyperparameters for each ablation experiment are the same as those used for the base model on each dataset.

\begin{itemize}
    \item \textbf{Single imputation block}: the model uses only one imputation block, specifically the forward block, avoiding unnecessary reverse operations. The Loss function is adjusted accordingly: the Masked Imputation Loss remains unchanged while the Observed Reconstruction Loss is reduced to its first term, as there is no second block to consider. The Discrepancy Loss is eliminated since only one block is used.
    
    \item \textbf{Weighted Combination}: instead of using the learned combining weights for the forward and backward representations, a simple average is used. Both representations are summed and divided by 2, following a similar approach to BRITS \cite{cao2018brits}. In this case, the original Loss function of the base model is retained.
\end{itemize}

Tables \ref{tab:physionet-ablation-part1} to \ref{tab:space-weather-ablation-part2} present the results of the ablation experiments for the studied datasets.

Tables \ref{tab:physionet-ablation-part1} and \ref{tab:physionet-ablation-part2} show the performance for the {\it PhysioNet} dataset. Among the three datasets, the ablation experiments here yield results closest to the complete model. Notably, the model without combining weights surpasses the complete model in the simplest configuration (MCAR with a 10\% missing rate). However, in all other cases, the base model achieves the best MAE scores. For the RMSE metric, both ablation experiments slightly outperform the complete model in the MCAR case with a 20\% missing rate, but their performance declines in more complex scenarios, where the full model demonstrates a greater advantage.

Tables \ref{tab:water-ablation-part1} and \ref{tab:water-ablation-part2} provide the evaluation results for the {\it Water Quality} dataset. Here, both ablation experiments perform similarly, with competitive results. However, the complete model consistently outperforms the ablation experiments across all configurations. Between the two ablation experiments, the model using both imputation blocks but without weighted combination performs better, particularly in the more complex scenarios involving sequences of random lengths.

Finally, Tables \ref{tab:space-weather-ablation-part1} and \ref{tab:space-weather-ablation-part2} present the metrics for the {\it Space Weather} dataset. The results show similar performance between the ablation models, especially in easier configurations. This indicates that the combining weights contribute less in simpler scenarios, such as the MCAR case. However, their impact becomes more pronounced in complex configurations, such as sequences of random lengths, where the complete model achieves the highest performance gains.

Although both ablation experiments demonstrate robust performance, the complete model outperforms them in nearly all cases for the MAE metric and in approximately 95\% of the cases for the RMSE metric. When the ablation models do slightly better, the differences are minor compared to the significant improvements seen with the complete model, particularly in challenging configurations. Nevertheless, the single-block model, with its significantly reduced number of parameters, may be appealing in scenarios where computational cost is a critical factor and the required imputation performance can be achieved with the smaller model.

\begin{table}[!ht]
    \centering
    \caption{Results of the ablation experiments for the {\it PhysioNet} dataset (Part 1). \mname\ (1B) represents the experiment with only one imputation block, and \mname\ (noWC) represents the experiment without the combining weights, using an average to combine both imputation blocks. \mname\ is the complete model.\label{tab:physionet-ablation-part1}}
    \renewcommand{\arraystretch}{1.2}    
    \begin{tabular}{|l|M{35pt}|M{35pt}|M{35pt}|M{35pt}|M{35pt}|M{35pt}|M{35pt}|M{35pt}|M{35pt}|}
    \hline
         \textbf{Physionet} & \multicolumn{3}{c|}{0.1 random} & \multicolumn{3}{c|}{0.2 random} & \multicolumn{3}{c|}{0.1 series 5} \\ \hline
        \textit{Metric} & MAE & RMSE & MRE & MAE & RMSE & MRE & MAE & RMSE & MRE \\ \hline
        \mname\ (1B) & 0.1890 & 0.4790 & 0.2712 & 0.1998 & 0.4387 & 0.2858 & 0.2524 & 0.4511 & 0.3487 \\ \hline
        \mname\ (noWC) & \textbf{0.1853} & \textbf{0.4658} & \textbf{0.2659} & 0.1978 & \textbf{0.4379} & 0.2829 & 0.2523 & 0.4496 & 0.3485 \\ \hline
        \mname\ & 0.1854 & 0.4786 & 0.2660 & \textbf{0.1974} & 0.4404 & \textbf{0.2823} & \textbf{0.2507} & \textbf{0.4493} & \textbf{0.3463} \\ \hline        
    \end{tabular}    
\end{table}

\begin{table}[!ht]
    \centering
    \caption{Results of the ablation experiments for the {\it PhysioNet} dataset (Part 2). \mname\ (1B) represents the experiment with only one imputation block, and \mname\ (noWC) represents the experiment without the combining weights, using an average to combine both imputation blocks. \mname\ is the complete model.\label{tab:physionet-ablation-part2}}
    \renewcommand{\arraystretch}{1.2}    
    \begin{tabular}{|l|M{35pt}|M{35pt}|M{35pt}|M{35pt}|M{35pt}|M{35pt}|M{35pt}|M{35pt}|M{35pt}|}
    \hline
         \textbf{Physionet} & \multicolumn{3}{c|}{0.2 series 5} & \multicolumn{3}{c|}{0.1 range 3-10} & \multicolumn{3}{c|}{0.2 range 3-10} \\ \hline
        \textit{Metric} & MAE & RMSE & MRE & MAE & RMSE & MRE & MAE & RMSE & MRE \\ \hline
        \mname\ (1B) & 0.2941 & 0.5113 & 0.4096 & 0.3477 & 0.6000 & 0.4925 & 0.4255 & \textbf{0.6900} & 0.6084 \\ \hline
        \mname\ (noWC) & 0.2992 & 0.5126 & 0.4168 & 0.3579 & 0.6086 & 0.5069 & 0.4272 & 0.6915 & 0.6109 \\ \hline
        \mname\ & \textbf{0.2925} & \textbf{0.5096} & \textbf{0.4074} & \textbf{0.3463} & \textbf{0.5999} & \textbf{0.4904} & \textbf{0.4193} & 0.6915 & \textbf{0.5996} \\ \hline        
    \end{tabular}    
\end{table}

\begin{table}[!ht]
    \centering
    \caption{Results of the ablation experiments for the {\it Water Quality} dataset (Part 1). \mname\ (1B) represents the experiment with only one imputation block, and \mname\ (noWC) represents the experiment without the combining weights, using an average to combine both imputation blocks. \mname\ is the complete model.\label{tab:water-ablation-part1}}
    \renewcommand{\arraystretch}{1.2}    
    \begin{tabular}{|l|M{30pt}|M{30pt}|M{30pt}|M{30pt}|M{30pt}|M{30pt}|M{30pt}|M{30pt}|M{30pt}|}
    \hline
         \textbf{Water Quality} & \multicolumn{3}{c|}{0.1 random} & \multicolumn{3}{c|}{0.2 random} & \multicolumn{3}{c|}{0.1 series 5} \\ \hline
        \textit{Metric} & MAE & RMSE & MRE & MAE & RMSE & MRE & MAE & RMSE & MRE \\ \hline
        \mname\ (1B) & 0.0248 & 0.1132 & 0.0491 & 0.0268 & 0.1254 & 0.0530 & 0.0295 & 0.1086 & 0.0581 \\ \hline
        \mname\ (noWC) & 0.0247 & 0.1146 & 0.0489 & 0.0246 & 0.1142 & 0.0487 & 0.0315 & 0.1323 & 0.0619 \\ \hline
        \mname\ & \textbf{0.0206} & \textbf{0.1046} & \textbf{0.0408} & \textbf{0.0216} & \textbf{0.1029} & \textbf{0.0427} & \textbf{0.0275} & \textbf{0.1004} & \textbf{0.0540} \\ \hline        
    \end{tabular}    
\end{table}

\begin{table}[!ht]
    \centering
    \caption{Results of the ablation experiments for the {\it Water Quality} dataset (Part 2). \mname\ (1B) represents the experiment with only one imputation block, and \mname\ (noWC) represents the experiment without the combining weights, using an average to combine both imputation blocks. \mname\ is the complete model.\label{tab:water-ablation-part2}}
    \renewcommand{\arraystretch}{1.2}    
    \begin{tabular}{|l|M{30pt}|M{30pt}|M{30pt}|M{30pt}|M{30pt}|M{30pt}|M{30pt}|M{30pt}|M{30pt}|}
    \hline
         \textbf{Water Quality} & \multicolumn{3}{c|}{0.2 series 5} & \multicolumn{3}{c|}{0.1 range 3-10} & \multicolumn{3}{c|}{0.2 range 3-10} \\ \hline
        \textit{Metric} & MAE & RMSE & MRE & MAE & RMSE & MRE & MAE & RMSE & MRE \\ \hline
        \mname\ (1B) & 0.0330 & 0.1335 & 0.0653 & 0.0497 & 0.1525 & 0.0984 & 0.0699 & 0.1749 & 0.1386 \\ \hline
        \mname\ (noWC) & 0.0334 & 0.1339 & 0.0660 & 0.0473 & 0.1443 & 0.0938 & 0.0635 & 0.1668 & 0.1258 \\ \hline
        \mname\ & \textbf{0.0278} & \textbf{0.0920} & \textbf{0.0551} & \textbf{0.0460} & \textbf{0.1385} & \textbf{0.0911} & \textbf{0.0629} & \textbf{0.1536} & \textbf{0.1246} \\ \hline        
    \end{tabular}    
\end{table}


\begin{table}[!ht]
    \centering
    \caption{Results of the ablation experiments for the {\it Space Weather} dataset (Part 1). \mname\ (1B) represents the experiment with only one imputation block, and \mname\ (noWC) represents the experiment without the combining weights, using an average to combine both imputation blocks. \mname\ is the complete model.\label{tab:space-weather-ablation-part1}}
    \renewcommand{\arraystretch}{1.2}    
    \begin{tabular}{|l|M{30pt}|M{30pt}|M{30pt}|M{30pt}|M{30pt}|M{30pt}|M{30pt}|M{30pt}|M{30pt}|}
    \hline
         \textbf{Space Weather} & \multicolumn{3}{c|}{0.1 random} & \multicolumn{3}{c|}{0.2 random} & \multicolumn{3}{c|}{0.1 series 5} \\ \hline
        \textit{Metric} & MAE & RMSE & MRE & MAE & RMSE & MRE & MAE & RMSE & MRE \\ \hline
        \mname\ (1B) & 0.0491 & 0.1273 & 0.0780 & 0.0540 & 0.1375 & 0.0857 & 0.0702 & 0.1916 & 0.1117 \\ \hline
        \mname\ (noWC) & 0.0482 & 0.1258 & 0.0765 & 0.0543 & 0.1356 & 0.0862 & 0.0687 & 0.1896 & 0.1094 \\ \hline
        \mname\ & \textbf{0.0471} & \textbf{0.1255} & \textbf{0.0748} & \textbf{0.0528} & \textbf{0.1354} & \textbf{0.0838} & \textbf{0.0681} & \textbf{0.1881} & \textbf{0.1083} \\ \hline        
    \end{tabular}    
\end{table}

\begin{table}[!ht]
    \centering
    \caption{Results of the ablation experiments for the {\it Space Weather} dataset (Part 2). \mname\ (1B) represents the experiment with only one imputation block, and \mname\ (noWC) represents the experiment without the combining weights, using an average to combine both imputation blocks. \mname\ is the complete model.\label{tab:space-weather-ablation-part2}}
    \renewcommand{\arraystretch}{1.2}    
    \begin{tabular}{|l|M{30pt}|M{30pt}|M{30pt}|M{30pt}|M{30pt}|M{30pt}|M{30pt}|M{30pt}|M{30pt}|}
    \hline
         \textbf{Space Weather} & \multicolumn{3}{c|}{0.2 series 5} & \multicolumn{3}{c|}{0.1 range 3-10} & \multicolumn{3}{c|}{0.2 range 3-10} \\ \hline
        \textit{Metric} & MAE & RMSE & MRE & MAE & RMSE & MRE & MAE & RMSE & MRE \\ \hline
        \mname\ (1B) & 0.0770 & 0.1959 & 0.1226 & 0.0802 & 0.2211 & 0.1277 & 0.0943 & 0.2358 & 0.1499 \\ \hline
        \mname\ (noWC) & 0.0759 & 0.1958 & 0.1208 & 0.0801 & 0.2189 & 0.1275 & 0.0933 & 0.2291 & 0.1482 \\ \hline
        \mname\ & \textbf{0.0754} & \textbf{0.1932} & \textbf{0.1200} & \textbf{0.0787} & \textbf{0.2119} & \textbf{0.1253} & \textbf{0.0919} & \textbf{0.2249} & \textbf{0.1460} \\ \hline        
    \end{tabular}    
\end{table}

\section{Analysis and Discussion of the Results}
The experimental results demonstrate the strengths of the proposed model, particularly when compared to other baseline methods and ablation variants. Below, we provide a critical analysis of why \mname\ achieves superior performance in the evaluated scenarios and the factors influencing its behavior.

\subsection{Role of Attention Mechanisms and Recurrent Layers}
\mname\ combines attention mechanisms with RNN-based layers, offering a unique advantage over both purely recurrent and self-attention-based architectures. The attention mechanisms dynamically prioritize relevant time steps, which is crucial for imputing long sequences or dealing with sparse data. Meanwhile, the GRU layers ensure the capture of local temporal dependencies.

For instance, in Tables  \ref{tab:water_part1} and \ref{tab:water_part2}, \mname\ consistently outperforms the baseline models across all configurations, particularly in scenarios with missing values in a sequence of random length. This advantage stems from the synergy between attention and recurrent mechanisms, allowing the model to better handle irregular gaps in the data by dynamically weighting relevant information.

\subsection{Performance in Missing Values in Sequences of Random-Length}
The random-length missing sequence configuration represents the most challenging scenario, as highlighted in Tables \ref{tab:physionet-ablation-part1} to \ref{tab:space-weather-ablation-part2}. The superior performance of \mname\ in these cases can be attributed to:
\begin{itemize}
    \item \textbf{Attention mechanisms}: these mechanisms allow the model to focus on distant, non-missing time steps, which are critical for imputing long gaps.
    \item \textbf{Bidirectional imputation}: by processing sequences in both forward and backward directions, \mname\ integrates information from both past and future time steps, providing a complete context.
\end{itemize}

Additionally, compared to models such as BRITS or SAITS, \mname\ exhibits more consistent performance in handling random-length sequences due to its ability to dynamically adapt to varying gaps in the data.

\subsection{Insights from Ablation Experiments}
The ablation experiments (Tables \ref{tab:physionet-ablation-part1} to \ref{tab:space-weather-ablation-part2}) provide valuable information on the contributions of each architectural component:
\begin{itemize}
    \item \textbf{Single Imputation Block}: removing one of the imputation blocks reduces the model’s capacity to process bidirectional dependencies. As a result, the model performs worse in scenarios requiring complex imputation, such as random-length sequences, as observed in Tables \ref{tab:physionet-ablation-part2} and \ref{tab:space-weather-ablation-part2}.
    \item \textbf{Weighted Combination}: replacing the learned combination weights with a simple average diminishes the model’s ability to dynamically prioritize forward or backward imputation. This limitation is evident in the {\it Space Weather} dataset (Tables \ref{tab:space-weather-ablation-part1} and \ref{tab:space-weather-ablation-part2}), where the complete model consistently outperforms this ablation variant in most configurations.
\end{itemize}

These experiments highlight that both components —the bidirectional structure and the learned combining weights— are critical for achieving SoA performance, particularly in complex scenarios.

\subsection{Comparison with Baseline Models}
The results from Tables \ref{tab:physionet_part1} to \ref{tab:space-weather-ablation-part2} show that \mname\ outperforms all baseline methods in almost every metric in all datasets. However, the advantages are more pronounced in the following cases:
\begin{itemize}
    \item \textbf{PhysioNet dataset}: due to its high sparsity, \mname\  ability to capture long- and short-term dependencies gives it a clear edge over models such as MRNN and BRITS, particularly in configurations with sequential missing values.
    \item \textbf{Water Quality dataset}: in scenarios with high missing rates, \mname's ability to dynamically adapt to irregular patterns results in consistently lower RMSE and MAE compared to SAITS or BRITS models. As shown in Tables \ref{tab:water_part1} and \ref{tab:water_part2}, the advantage of \mname\ is particularly evident in random-length missing sequences, where other models struggle to converge effectively.
    \item \textbf{Space Weather dataset}: the performance gap is narrower in simpler configurations, such as MCAR, as shown in Tables \ref{tab:space-weather-ablation-part1} and \ref{tab:space-weather-ablation-part2}. However, in more complex scenarios involving random-length sequences, \mname\ achieves significantly better results, showcasing its robustness in handling intricate temporal patterns.
\end{itemize}

\subsection{Relevance of the Space Weather Dataset}
The {\it Space Weather} dataset, although less commonly used compared to {\it PhysioNet} or {\it Water Quality} datasets, represents a critical domain for analyzing complex time series. Missing data in this context often arises in satellite observations, where gaps can severely impact predictions of {\it Space Weather} phenomena, such as solar storms or geomagnetic disturbances. Accurate imputation in this domain is essential for improving models used to forecast disruptions to satellite navigation, telecommunications, or power grid infrastructure. The results show that \mname\  performs well in handling irregular and challenging missing patterns, making it a suitable candidate for such high-stakes applications.

\section{Conclusions}
\label{sec:conclusions}
Time series are widely used and hold significant value in multiple data analysis applications. However, in most real-world scenarios, data is often imperfect, and the presence of missing values is a common occurrence that adversely impacts downstream applications.  
The causes of missing data are diverse and can follow various patterns. For instance, missing values may occur at random, with no discernible relationship, or they may be linked to malfunctions or interruptions in the communication systems of measurement devices. In such cases, missing values are likely to appear in sequences of arbitrary length.

Building on previous RNN and Self-Attention models, we proposed \mname, a novel approach for time series imputation that combined the strengths of both techniques. \mname\ was evaluated on three datasets with three distinct missing data patterns: missing values completely at random (MCAR), missing values in sequences of fixed length, and missing values in sequences of variable length. Across all datasets and configurations, the proposed model consistently outperformed the baseline models, demonstrating its robustness and adaptability.

Another contribution of this work is the introduction of the {\it Space Weather} dataset, a realistic benchmark derived from space probe observations. By capturing complex missing data patterns, including Missing Not At Random (MNAR) scenarios, this dataset provides a valuable resource for the Deep Learning community. We hope its availability fosters contributions from the community, advancing research and innovation in the domain.



\section*{Acknowledgements}

Armando Collado-Villaverde is supported by the ESA project \textit{Deep Neural Networks for Geomagnetic Forecasting} 4000137421/22/NL/GLC/my.

\bibliographystyle{alpha}
\bibliography{brati}

\end{document}